# Stack-Captioning: Coarse-to-Fine Learning for Image Captioning

Jiuxiang Gu[1], Jianfei Cai[2], Gang Wang[3], Tsuhan Chen[2]

[1] ROSE Lab, Interdisciplinary Graduate School, Nanyang Technological University, Singapore
[2] School of Computer Science and Engineering, Nanyang Technological University, Singapore
[3] Alibaba AI Labs, Hangzhou, China
{jgu004, asjfcai, tsuhan}@ntu.edu.sg, gangwang6@gmail.com

## Abstract

The existing image captioning approaches typically train a one-stage sentence decoder, which is difficult to generate rich fine-grained descriptions. On the other hand, multi-stage image caption model is hard to train due to the vanishing gradient problem. In this paper, we propose a coarse-to-fine multi-stage prediction framework for image captioning, composed of multiple decoders each of which operates on the output of the previous stage, producing increasingly refined image descriptions. Our proposed learning approach addresses the difficulty of vanishing gradients during training by providing a learning objective function that enforces intermediate supervisions. Particularly, we optimize our model with a reinforcement learning approach which utilizes the output of each intermediate decoder's test-time inference algorithm as well as the output of its preceding decoder to normalize the rewards, which simultaneously solves the well-known exposure bias problem and the loss-evaluation mismatch problem. We extensively evaluate the proposed approach on MSCOCO and show that our approach can achieve the state-of-the-art performance.

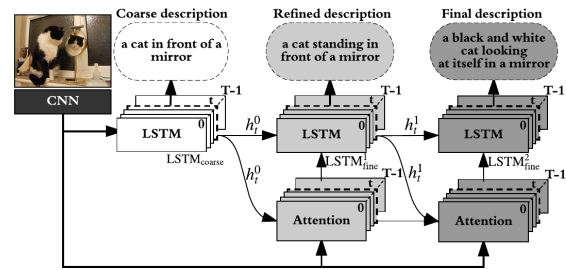

Figure 1: Illustration of our proposed coarse-to-fine framework. Our model consists of one image encoder (CNN) and a sequence of sentence decoders (attention-based LSTM networks), and it takes the image as input and refines the image descriptions from coarse to fine. Here we show the increasingly improved image descriptions in two stages (gray and dark gray).

## Introduction

The challenge of image captioning lies in designing a model that can effectively utilize the image information and generate more human-like rich image descriptions. Motivated by the recent advances in natural language processing, current image captioning approaches typically follow the encoding-decoding framework (Ranzato et al. 2016), which consists of a Convolutional Neural Network (CNN) based image encoder and a Recurrent Neural Network (RNN) based sentence decoder, with various variants for image captioning (Fang et al. 2015; Mao et al. 2014; Wu et al. 2016). Most of these existing image captioning approaches are trained by maximizing the likelihood of each ground-truth word given the previous ground-truth words and the image with back propagation.

There are three major problems in these existing image captioning methods. Firstly, it is extremely hard for them to generate rich fine-grained descriptions. This is because rich descriptions require high-complexity models, where the problem of vanishing gradients often occurs, considering the back-propagated gradients diminish in strength as they propagate through many layers of a complex network. Secondly, there is an exposure bias between the training and the testing (Ranzato et al. 2016; Wiseman and Rush 2016; Gu, Cho, and Li 2017). Specifically, the sentence decoder is trained to predict a word given the previous ground-truth words, while at testing time, the caption generation is accomplished by greedy search or with beam search, which predicts the next word based on the previously generated words that is different from the training mode. Since the model has never been exposed to its own predictions, it will result in error accumulation at test time. To address the exposure bias problem, scheduled sampling (Bengio et al. 2015), *i.e.*, randomly selecting between previous ground-truth words and previously generated words, has become the current dominant training procedure to fit RNNs based models. However, it can only mitigate the exposure bias but cannot largely solve it. Thirdly, there is a loss-evaluation mismatch (Ranzato et al. 2016). Specifically, language models are usually trained to minimize the cross-entropy loss at each time-step, while at testing time, we evaluate the generated captions with the sentence-level evaluation met-rics, *e.g.*, BLEU-$n$ (Papineni et al. 2002), CIDEr (Vedantam, Lawrence Zitnick, and Parikh 2015), SPICE (Anderson et al. 2016), *etc.*, which are non-differentiable and cannot be directly used as training loss.

In this paper, considering the great challenge of generating rich image descriptions in one stage, we propose a coarse-to-fine multi-stage prediction framework. Our model consists of an image encoder and a sequence of sentence decoders that repeatedly generate image descriptions in finer details. However, directly composing such multi-stage decoders in an image captioning model faces the risk of the vanishing gradients problem. Motivated by the works on image recognition (Zhang, Lee, and Lee 2016; Fu, Zheng, and Mei 2017), which show that supervising very deep networks at intermediate layers aids in learning, we also enforce intermediate supervisions for each stage. Furthermore, inspired by the recent image captioning work (Rennie et al. 2017), which uses Reinforcement Learning (RL) to address the loss-evaluation mismatch problem and include the inference process as a baseline in training to address the exposure bias problem, we also design a similar RL-based training method but extend it from one-stage (Rennie et al. 2017) to our multi-stage framework, where rewards are introduced at each stage as intermediate supervision. Particularly, we optimize our model with a RL-based approach which utilizes the output of each intermediate decoder's test-time inference algorithm as well as the output of its preceding decoder to normalize the rewards. In addition, to cope with our coarse-to-fine learning framework, we adopt a stacked attention model to extract more fine-grained visual attention information for word prediction at each stage. Figure 1 illustrates our proposed coarse-to-fine framework, which consists of three stacked Long Short-Term Memory (LSTM) networks. The first LSTM generates the coarse-scale image description, and the subsequent LSTM networks serve as the fine-scale decoders. At each stage in our model, attention weights and hidden vector produced by the preceding stage are used as inputs, which are taken as the disambiguating cues to the subsequent stage. As a result, each stage of the decoder generates words with increasingly refined attention weights as well as words.

The main contributions of this work include: (a) a coarse-to-fine framework which increases the model complexity gradually with increasingly refined attention weights for image captioning and (b) a reinforcement learning method that directly optimizes model with the normalized intermediate rewards. Experiments show outstanding performance of our approach on MSCOCO (Lin et al. 2014).

# Related Works

**Image Captioning with Maximum Likelihood Estimation.** The information gap between the visual content of the images and their corresponding descriptions has been extensively studied (Vinyals et al. 2015; Fang et al. 2015; Mao et al. 2014; Wu et al. 2016). The classical image captioning framework is based on the CNN image encoder and the RNN based sentence decoder (Vinyals et al. 2015). Only providing the global image feature is not sufficient, as the power of RNNs lies in its capability to model the contextual information between time steps, while the global image representation weakens the RNN's memory of the visual information. To better incorporate the image information into the language processing, a few approaches have been proposed (You et al. 2016; Yang et al. 2016b). Visual attention for image captioning was first being introduced by (Xu et al. 2015) which incorporates the spatial attention on convolutional features of images into the encoder-decoder framework through the *soft* and *hard* attention mechanisms. Their work was later followed by (Yang et al. 2016a) and (Liu et al. 2017b) which further improves the visual attention mechanism. However, all these approaches are typically trained by maximising the likelihood estimation, often called as *Teacher-Forcing* (Williams and Zipser 1989). Instead of training the model with the handcrafted loss, some researchers applied the adversarial training for image captioning, called *Professor-Forcing* (Lamb et al. 2016), which uses adversarial training to encourage the dynamics of the RNNs to be the same as that of training conditioned on previous ground truth words.

Recently, some works have proposed to encode more discriminative visual information into the captioning model. They leverage visual attributes of the image to enhance the visual information using some weakly supervised approach. In (You et al. 2016; Yao et al. 2017), they incorporate high-level attributes into the encoder-decoder framework and achieve large improvements. Both of (You et al. 2016) and (Wu et al. 2016) treat the attribute detection problem as a multi-instance learning (MIL) problem and train a corresponding CNN by minimizing the element-wise logistic loss function. (Liu et al. 2017a) uses R-FCN (Li et al. 2016) to detect the visual attributes and adopts a sequential attention mechanism to translate the attributes to a word sequence.

**Image Captioning with Reinforcement Learning.** Several attempts have been made to use reinforcement learning to address the discrepancy between the training and the testing objectives for image captioning (Rennie et al. 2017). The first work of training RNN-based sequence model with policy gradient was proposed by (Ranzato et al. 2016), in which a REINFORCE-based approach was used to calculate the sentence-level reward and a Monte-Carlo technique was employed for training. Similarly, (Liu et al. 2017c) estimates the action value by averaging three roll-out sequences which is the same as (Yu et al. 2017). Instead of using the sentence-level reward in training, (Bahdanau et al. 2017) use the token-level reward in temporal difference training for sequence generation. Recently, the *self-critical learning* approach proposed by (Rennie et al. 2017) utilizes an improved REINFORCE algorithm with a reward obtained by the current model against the baseline, *i.e.*, the inference algorithm.

All these existing researches on image captioning mainly focus on one-stage training (Mao et al. 2014; Vinyals et al. 2015; Rennie et al. 2017). However, it is challenging to generate a rich description for the image in one stage. Rather than generating image description in one-step, in this paper, we propose a coarse-to-fine model by stacking multiple intermediate sentence decoders and optimizing them with sentence-level evaluation metrics, where the coarse decoder generates the coarse caption and reduces the computational burden for the fine-scale sentence decoders to generate complex and rich image descriptions. Note that our coarse-to-fine concept at high level is similar to the coarse-to-fine reasoning (Kiddon and Domingos 2011), while the latter is not for image captioning. Our RL-based supervision for solving the loss-evaluation mismatch problem is related to (Rennie et al. 2017), while ours is designed for our multi-stage coarse-to-fine model and (Rennie et al. 2017) is for the conventional one-stage model.

# Methodology

In this paper, we consider the problem of learning to generate image description $\hat{\mathbf{Y}} = \{\hat{Y}_0, \ldots, \hat{Y}_{T-1}\}$ for an image $\mathbf{I}$, where $\hat{Y}_t \in \mathcal{D}$ is the predicted word, $\mathcal{D}$ is the dictionary, and $T$ denotes the sequence length. Our algorithm builds a coarse-to-fine model with the same target as those one-stage models, but with the additional intermediate layers between the output layer and the input layer. We first train the model by maximizing log-likelihood of each successive target word conditioned on the input image and the gold history of target words $\mathbf{Y} = \{Y_0, \ldots, Y_{T-1}\}$, and then optimize the model with sentence-level evaluation metrics. We denote by $\hat{\mathbf{Y}}^i, i \in \{0, \cdots, N_f\}$ the predicted word sequence of the $i$-th stage decoder, and $N_f$ is the total number of fine stages. As a result, each intermediate sentence decoder predicts the increasingly refined image description, and the prediction of the last decoder is taken as the final image description. Note that we treat stage $i = 0$ as the coarse decoder, and stages $i >= 1$ as the fine decoders.

## Image Encoding

We first encode the given image $\mathbf{I}$ to the spatial image features $\mathbf{V} = \{V_0, \cdots, V_{k \times k - 1}\}$, $V_i \in \mathbb{R}^{d_v}$ with CNN: $\mathbf{V} = \text{CNN}(\mathbf{I})$, where $k \times k$ is the number of regions, each feature channel $V_i$ depicts a region of the image, and $d_v$ is the dimension of the feature vector for each region. Specifically, we extract the image features from the final convolutional layer of CNN, and use spatial adaptive average pooling to resize the features to a fixed-size spatial representation of $k \times k \times d_v$.

## Coarse-to-Fine Decoding

The overall coarse-to-fine sentence decoder consists of one coarse decoder and a sequence of attention-based fine decoders that repeatedly produce refined attention maps for the prediction of each word based on the cues from the preceding decoder. The first stage of our model is a coarse decoder which predicts coarse description from the global image feature. In the subsequent stages, each stage $i \in \{1, \cdots, N_f\}$ is a fine decoder which predicts the improved image description based on image features and the outputs of the preceding stage. Particularly, we use the attention weights of the preceding stage to provide the following stage beliefs of regions for word prediction. More formally, we decode the image features in multiple stages, where the prediction $\hat{\mathbf{Y}}^i$ of each stage is a refinement of the prediction $\hat{\mathbf{Y}}^{i-1}$ of previous stage.

Figure 2 illustrates the coarse-to-fine decoding architecture, where the top row contains one coarse decoder and two stacked attention-based fine decoders under the training mode, and the bottom row shows the fine decoders under its inference mode (greedy decoding) for computing rewards so as to incorporate intermediate supervisions. In the following, we will introduce the adopted coarse decoder, our proposed fine decoder, our proposed stacked attention model and our proposed RL-based process for incorporating intermediate supervisions.

**Coarse Decoder.** We start by decoding in a coarse search space in the first stage ($i = 0$), where we learn a coarse decoder with an LSTM network, called LSTM$_{\text{coarse}}$. At each time step $t \in [0, T-1]$, the input to LSTM$_{\text{coarse}}$ consists of the previous target word $y_{t-1}$, concatenated with the global image feature, and the previous hidden states. The operation of the LSTM$_{\text{coarse}}$ can be described as:

$$o_t^0, h_t^0 = \text{LSTM}_{\text{coarse}}(h_{t-1}^0, i_t^0, y_{t-1}) \quad (1)$$

$$i_t^0 = [f(\mathbf{V}); h_{t-1}^{N_f}] \quad (2)$$

where $h_{t-1}^0$ and $h_{t-1}^{N_f}$ are the hidden states, $o_t^0$ is the cell output, $y_{t-1} = \mathbf{W}_e Y_{t-1}$ is the embedding of previous word $Y_{t-1}$. We obtain the global image feature $f(\mathbf{V})$ by taking a mean-pooling over the spatial image features as $\frac{1}{k \times k} \sum_{i=0}^{k \times k - 1} V_i$. The $t$-th decoded word $\hat{Y}_t^0$ of LSTM$_{\text{coarse}}$ is drawn from the dictionary $\mathcal{D}$ according to the softmax probability: $\hat{Y}_t^0 \sim \text{Softmax}(\mathbf{W}_o^0 o_t^0 + \mathbf{b}_o^0)$.

**Fine Decoder.** In the subsequent stages, each fine decoder predicts the word $\hat{Y}_t^i$ based on the image features $\mathbf{V}$ again, and the attention weights $\boldsymbol{\alpha}_t^{i-1}$ and the hidden state $h_t^{i-1}$ from the preceding LSTM. Each fine decoder consists of an LSTM$_{\text{fine}}^i$ network and an attention model. At each time step $t$, the input to LSTM$_{\text{fine}}^i$ consists of the attended image feature, the previous word embedding $y_{t-1}$, its previous hidden state $h_{t-1}^i$, and the updated hidden state $h_t^{i-1}$ from the preceding LSTM. Note that when $t = 1$, $h_t^0$ is the hidden output of LSTM$_{\text{coarse}}$; otherwise $h_t^{i-1}$ is the hidden output of the preceding LSTM$_{\text{fine}}^{i-1}$. Therefore, the updating procedure of LSTM$_{\text{fine}}^i$ can be written as:

$$o_t^i, h_t^i = \text{LSTM}_{\text{fine}}^i(h_{t-1}^i, i_t^i, y_{t-1}) \quad (3)$$

$$i_t^i = [g(\mathbf{V}, \boldsymbol{\alpha}_t^{i-1}, h_t^{i-1}); h_t^{i-1}] \quad (4)$$

where $o_t^i$ is the cell output of LSTM$_{\text{fine}}^i$, and $g(\cdot)$ is the spatial attention function which feeds attended visual representations as the additional inputs to LSTM$_{\text{fine}}^i$ at each time step to emphasise the detailed visual information. During the inference, the final output word $\hat{Y}_t$ is drawn from $\mathcal{D}$ according to the softmax probability: $\hat{Y}_t \sim \text{Softmax}(\mathbf{W}_o^{N_f} o_t^{N_f} + \mathbf{b}_o^{N_f})$.

**Stacked Attention Model.** As aforementioned, our coarse decoder generates words based on the global image features. However, in many cases, each word is only related to a small region of an image. Using the global image feature for word prediction could lead to sub-optimal results due to the noises introduced from the irrelevant regions for each prediction (Gu et al. 2017b).

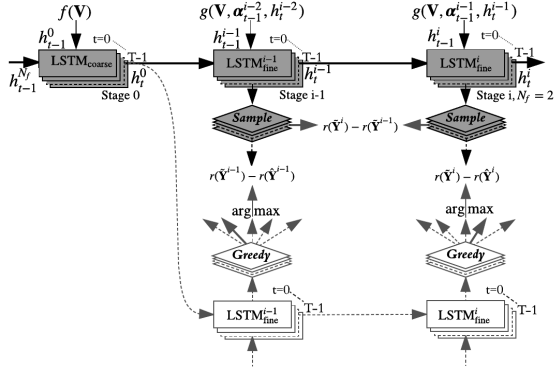

Figure 2: Illustration of the proposed coarse-to-fine decoding using intermediate supervision (reward) after each stage. The top row (gray) contains one coarse decoder (left) and two visual attention-based fine decoders under the training mode. The bottom row shows the fine decoders under its inference mode (greedy decoding) for computing rewards.

Therefore, the attention mechanism has been introduced to significantly improve the performance of image captioning. It typically produces a spatial map highlighting image regions relevant to each predicted word. In this research, to extract more fine-grained visual information for word prediction, we adopt a stacked attention model to filter out noises gradually and pinpoint the regions that are highly relevant to the word prediction. In each fine stage $i$, our attention model operates on both image features $\mathbf{V}$ and attention weights $\boldsymbol{\alpha}_t^{i-1}$ from the preceding stage.

Formally, for the time step $t$ of stage $i$, the stacked attention model is defined as:

$$g(\mathbf{V}, \boldsymbol{\alpha}_t^{i-1}, h_t^{i-1}) = \sum_{n=0}^{k \times k - 1} \alpha_t^{i,n} \cdot (\mathbf{W}_{v\alpha}^i V_n + \mathbf{b}_{v\alpha}^i) \quad (5)$$

where $\alpha_t^{i,n}$ corresponds to the attention probability of each image region. We compute the attention probability $\alpha_t^{i,n}$ as follows:

$$\boldsymbol{\alpha}_t^i = \text{softmax}(\mathbf{W}_\alpha^i \boldsymbol{A}_t^i + \mathbf{b}_\alpha^i) \quad (6)$$

$$A_t^{i,n} = \tanh(\mathbf{W}_{va}^i V_n + \mathbf{W}_{ha}^i \bar{h}_t^{i-1}) \quad (7)$$

$$\bar{h}_t^{i-1} = h_t^{i-1} + \sum_{n=0}^{k \times k - 1} \alpha_t^{i-1,n} \cdot (\mathbf{W}_{v\alpha}^{i-1} V_n + \mathbf{b}_{v\alpha}^{i-1}) \quad (8)$$

where $h_t^{i-1}$ is the updated hidden state of $\text{LSTM}_{\text{fine}}^{i-1}$, which is added to the aggregated image features to form a new hidden representation $\bar{h}_t^{i-1}$. Note that when $i = 1$, we set $\boldsymbol{\alpha}_t^0$ to zero.

### Learning

The coarse-to-fine approach described above results in a deep architecture. Training such a deep network can be prone to the vanishing gradient problem, where the magnitude of gradients decreases in strength when back-propagated through multiple intermediate layers. A natural approach to address this problem is to incorporate supervised training objectives into the intermediate layers. Each stage of the coarse-to-fine sentence decoder is trained to predict the words repeatedly. We first train the network by defining a loss function for each stage $i$ that minimizes the cross-entropy (XE) loss, i.e.,

$$\mathcal{L}_{\text{XE}}^i(\theta_{0:i}) = -\sum_{t=0}^{T-1} \log(p_{\theta_{0:i}}(Y_t \mid Y_{0:t-1}, \mathbf{I})), \quad (9)$$

where the $Y_t$ is the ground-truth word, and $\theta_{0:i}$ is the parameters up to the stage-$i$ decoder. By adding the losses at each stage $i$, we obtain the overall learning objective for the full architecture:

$$\mathcal{L}_{\text{XE}}(\theta) = \sum_{i=0}^{N_f} \mathcal{L}_{\text{XE}}^i(\theta_{0:i})$$

$$= -\sum_{i=0}^{N_f} \sum_{t=0}^{T-1} \log(p_{\theta_{0:i}}(Y_t \mid Y_{0:t-1}, \mathbf{I})) \quad (10)$$

where $p_{\theta_{0:i}}(Y_t \mid Y_{0:t-1}, \mathbf{I})$ is the output probability of word $Y_t$ given by the $\text{LTSM}^i$ decoder. We share the weights of the models across all time steps.

However, training with the loss function of Equation 10 is not sufficient. As mentioned in Section 1, the existing log-likelihood training methods have the problem of the discrepancy between their training and testing modes, where the model is often trained with scheduled sampling, while in testing, greed decoding or beam search is commonly used to get higher scores. Besides, the log-likelihood score of the prediction does not correlate well with the standard evaluation metrics such as BLEU, and CIDEr. Many researchers have explored in the direction of optimizing the image captioning model with the evaluation metrics (e.g., CIDEr in (Rennie et al. 2017)). To optimise the evaluation metrics during each stage, we consider the image caption generation process as a reinforcement learning problem, i.e., given an environment (previous states), we want to get an agent (e.g., RNN, LSTM or GRU) to look at the environment (image features, hidden states, and previous words), and make an action (the prediction of the next word). After generating a complete sentence, the agent will observe a sentence-level reward and update its internal state.

We cast our generative model in the reinforcement learning terminology as in (Ranzato et al. 2016; Rennie et al. 2017). The LSTM-based decoder of each stage can be viewed as an agent that interacts with the external environment. The policy network parametrized by $\theta_{0:i}$ defines a policy $p_{\theta_{0:i}}$, which receives a state (preceding outputs, internal state of LSTM and image features) and produces an action $\tilde{Y}_t^i \sim p_{\theta_{0:i}}$ which is the prediction of the next word sampled from the LSTM at time step $t$. Once we have a complete predicted sentence $\tilde{\mathbf{Y}}^i$, the agent observes a reward $r(\tilde{\mathbf{Y}}^i)$ (*e.g.*, CIDEr score) of the sentence. The goal of RL-based training is to minimize the negative expected rewards (punishments) of multi-stages, :

$$\mathcal{L}_{\text{RL}}(\theta) = -\sum_{i=1}^{N_f} \mathbb{E}_{\tilde{\mathbf{Y}}^i \sim p_{\theta_{0:i}}}[r(\tilde{\mathbf{Y}}^i)] \approx -\sum_{i=1}^{N_f} r(\tilde{\mathbf{Y}}^i) \quad (11)$$

where $\tilde{\mathbf{Y}}^i = \{\tilde{Y}_0^i, \cdots, \tilde{Y}_{T-1}^i\}$, and $\tilde{Y}_t^i$ is sampled from the stage $i$ at time step $t$. $r(\tilde{\mathbf{Y}}^i)$ is calculated by comparing the generated sentence to the corresponding reference sentences using the standard evaluation metric. Note that we do not consider $i = 0$ in Equation 11 as the coarse decoder does not has a preceding stage. After that, we calculate the expected gradient using the Monte-Carlo sample $\tilde{\mathbf{Y}}^i$ from $p_{\theta_{0:i}}$ as:

$$\nabla_\theta \mathcal{L}_{\text{RL}}(\theta) = \sum_{i=1}^{N_f} \nabla_{\theta_{0:i}} \mathcal{L}_{\text{RL}}(\theta_{0:i}) \quad (12)$$

$$\approx -\sum_{i=1}^{N_f} r(\tilde{\mathbf{Y}}^i) \cdot \nabla_{\theta_{0:i}} \log p_{\theta_{0:i}}(\tilde{\mathbf{Y}}^i) \quad (13)$$

To reduce the variance of the gradient estimate in Equation 13, we follow the REINFORCE approach from SCST (Rennie et al. 2017) to approximate Equation 13 as:

$$\nabla_\theta \mathcal{L}_{\text{RL}}(\theta) \approx -\sum_{i=1}^{N_f} \Delta r(\tilde{\mathbf{Y}}^i) \cdot \nabla_{\theta_{0:i}} \log p_{\theta_{0:i}}(\tilde{\mathbf{Y}}^i) \quad (14)$$

where $\Delta r(\tilde{\mathbf{Y}}^i)$ is the relative reward which can reduce the variance of the gradient estimate. The principal idea of our RL-based coarse-to-fine learning approach is to baseline the REINFORCE algorithm with the reward $r(\hat{\mathbf{Y}}^i)$ obtained in each stage under the inference algorithm at test time, as well as the reward $r(\tilde{\mathbf{Y}}^{i-1})$ obtained by its preceding decoder at train time. Particularly, $\Delta r(\tilde{\mathbf{Y}}^i)$ is defined as:

$$\Delta r(\tilde{\mathbf{Y}}^i) = \left[r(\tilde{\mathbf{Y}}^i) - r(\hat{\mathbf{Y}}^i)\right] + \left[r(\tilde{\mathbf{Y}}^i) - r(\tilde{\mathbf{Y}}^{i-1})\right] \quad (15)$$

where $\tilde{\mathbf{Y}}^i$ is a sampled caption of the $i$-th stage and $\hat{\mathbf{Y}}^i$ is obtained by the conventional greedy decoding. The first term in Equation 15 tends to increase the probability of the samples of stage $i$ that score higher than the results of stage $i$ at test-mode (greedy decoding). In other words, we supress those samples that have the worse scores than the greedy decoding results. The second term increases the probability of the samples from stage $i$ that outperform the samples from stage $i - 1$, and suppresses the inferior samples.

# Experiments

In this section, we first describe the dataset used in our experiments, and then introduce the baseline methods for comparisons and the implementation details followed by the detailed results. We report all the results using MSCOCO caption evaluation tool[1].

[1] https://github.com/tylin/coco-caption

### Datasets and Setting
We evaluate the proposed approach on MSCOCO dataset. The dataset contrains 123,000 images, where each image has five reference captions. We follow the setting of (Karpathy and Fei-Fei 2015) by using 5,000 images for offline validation and 5,000 images for offline testing. The widely used BLEU, METEOR, ROUGE, CIDEr, and SPICE scores are used to measure the quality of the generated captions. We further test on the MSCOCO test set consisting of 40,775 images, and then conduct the online comparison against the state-of-the-art via the online MSCOCO evaluation server.

### Baseline Approaches for Comparisons
To gain insight into the effectiveness of our proposed approach, we compare the following models with each other:
**LSTM and LSTM$_{\text{3 layers}}$**. We implement a one layer LSTM-based image captioning model based on the framework proposed by (Vinyals et al. 2015). We also add two additional LSTM networks after the one layer LSTM model, which is named as LSTM$_{\text{3 layers}}$. We first train these two models with XE loss, and then optimize the CIDEr metric with SCST (Rennie et al. 2017).
**LSTM+ATT$_{\text{Soft}}$ and LSTM+ATT$_{\text{Top-Down}}$**. We implement two types of visual attention-based image captioning models: the Soft-attention model (LSTM+ATT$_{\text{Soft}}$) proposed by (Xu et al. 2015) and the Top-Down attention model (LSTM+ATT$_{\text{Top-Down}}$) proposed by (Anderson et al. 2017). We encode the image with ResNet-101 and apply the spatially adaptive pooling to get a fixed-size output of $14 \times 14 \times 2048$. At each time step, the attention model produces an attention mask over the 196 spatial locations. LSTM+ATT$_{\text{Top-Down}}$ consists of two LSTM networks, where the first LSTM takes the mean-pooled image feature as input, and the second LSTM predicts the words based on the attended image features and the hidden state of the first LSTM. Similarly, we also train these two models with XE Loss and the RL-based sentence-level metric.
**Stack-Cap and Stack-Cap\***. Stack-Cap is our proposed method and Stack-Cap\* is a simplified version. In particular, Stack-Cap\* incorporates the multiple attention models into LSTM$_{\text{3 layers}}$. Here we treat the first LSTM as the coarse decoder, and the subsequent two attention-based LSTM networks ($N_f = 2$) as the fine decoders. Stack-Cap has the architecture similar to Stack-Cap\*, except that it applies the proposed stacked attention model instead of the independent attention model. We train these two models (Stack-Cap\* and Stack-Cap) with the proposed coarse-to-fine (C2F) learning approach.

### Implementation Details
In this paper, we set the number of hidden units of each LSTM to 512, the number of hidden units in the attention layer to 512, and the vocabulary size of the word embedding to 9,487. In our implementation, the parameters are randomly initialized except the image CNN, for which we encode the full image with the ResNet-101 pre-trained on ImageNet.

We first train our model under the cross-entropy cost using Adam (Kingma and Ba 2015) optimizer with an initial learning rate of $4 \times 10^{-4}$ and a momentum parameter of 0.9. After that, we run the proposed RL-based approach on the just trained model to be optimized for the CIDEr metric. During this stage, we use Adam with a learning rate $5 \times 10^{-5}$. After each epoch, we evaluate the model on the validation set and select the model with the best CIDEr score for testing. During testing, we apply beam search which can increase the performance of greedy decoding. Unlike greedy decoding which keeps only a single hypothesis during decoding, Beam search keeps K > 1 (K = 5 in our experiments) hypotheses that have the highest scores at each time step, and returns the hypothesis with the highest log probability at the end.

## Quantitative Analysis

In this experiment, we first optimize the models with the standard cross-entropy (XE) loss. We report the performance of our model and the baselines on the Karpathy test split in Table 1. Note that all results are reported without fine-tuning of the ResNet-101.

| Approach | B@1 | B@2 | B@3 | B@4 | M | C |
|---|---|---|---|---|---|---|
| LSTM (XE) | 72.1 | 54.8 | 39.6 | 28.5 | 24.3 | 91.4 |
| LSTM$_{3\ layers}$ (XE) | 70.5 | 53.1 | 38.9 | 28.3 | 23.2 | 85.7 |
| LSTM+Att$_{Soft}$ (XE) | 73.8 | 57.2 | 43.1 | 33.0 | 25.7 | 101.0 |
| LSTM+Att$_{Top-Down}$ (XE) | 74.9 | 58.6 | 44.5 | 33.3 | 25.8 | 103.4 |
| Stack-Cap* (XE) | 75.6 | 59.6 | 45.6 | 34.6 | 26.3 | 108.0 |
| Stack-Cap (XE) | **76.2** | **60.4** | **46.4** | **35.2** | **26.5** | **109.1** |

Table 1: Performance comparisons on MSCOCO, where B@n is short for BLEU-n, M is short for METEOR, and C is short for CIDEr. All values are reported as the percentage (**Bold** numbers are the best results).

It can be seen from Table 1 that our coarse-to-fine image captioning model (Stack-Cap) achieves the best performances in all metrics. The two coarse-to-fine models, Stack-Cap and Stack-Cap*, give similar performance. Note that although these two coarse-to-fine models have the same number of LSTM units as LSTM$_{3\ layers}$, directly adding two additional LSTM layers in LSTM$_{3\ layers}$ without intermediate supervision decreases the performance of LSTM as the model experiences overfitting. Our coarse-to-fine approach can optimize the network gradually with the intermediate supervision and avoid overfitting. We also observe that Soft attention (LSTM+ATT$_{Soft}$) and Top-Down attention (LSTM+ATT$_{Top-Down}$) can significantly improve the performance of image captioning. Our best model (Stack-Cap) with stacked attention networks outperforms the Stack-Cap*, which demonstrates that adjusting the attention on the relevant visual clues progressively can generate better image descriptions.

After optimizing the models with XE loss, we optimize them for the CIDEr metric with the RL-based algorithms. The performances of the four models optimized for CIDEr with the SCST (Rennie et al. 2017) and the performances of two models optimized with the proposed coarse-to-fine (C2F) learning are also reported in Table 2. We can see

| Approach | B@1 | B@2 | B@3 | B@4 | M | C |
|---|---|---|---|---|---|---|
| LSTM (CIDEr) | 76.7 | 58.3 | 42.8 | 30.8 | 25.5 | 100.2 |
| LSTM$_{3\ layers}$ (CIDEr) | 73.0 | 56.1 | 41.1 | 29.9 | 25.1 | 95.9 |
| LSTM+Att$_{Soft}$ (CIDEr) | 77.3 | 59.3 | 44.1 | 32.1 | 25.9 | 104.8 |
| LSTM+Att$_{Top-Down}$ (CIDEr) | 76.7 | 60.4 | 45.6 | 33.9 | 26.5 | 112.7 |
| Stack-Cap* (C2F) | 77.9 | 61.6 | 46.7 | 35.0 | 26.9 | 115.9 |
| Stack-Cap (C2F) | **78.6** | **62.5** | **47.9** | **36.1** | **27.4** | **120.4** |

Table 2: Performance comparisons with the baselines on MSCOCO Karpathy test split. Our Stack-Cap (C2F) model achieves significant grains across all metrics.

that our Stack-Cap model obtains significant gains across all metrics.

Table 3 compares the results of our Stack-Cap (C2F) model with those of the existing methods on MSCOCO Karpathy test split, where Stack-Cap achieves the best performance in all metrics.

**Online Evaluation.** Table 4 reports the performance of our proposed Stack-Cap model trained with the coarse-to-fine learning on the official MSCOCO evaluation server[2]. We can see that our approach achieves very competitive performance, compared to the state-of-the-art. Note that the results of SCST:Att2in (Ens. 4) are achieved by the ensemble of four models, while our results are generated by the single model.

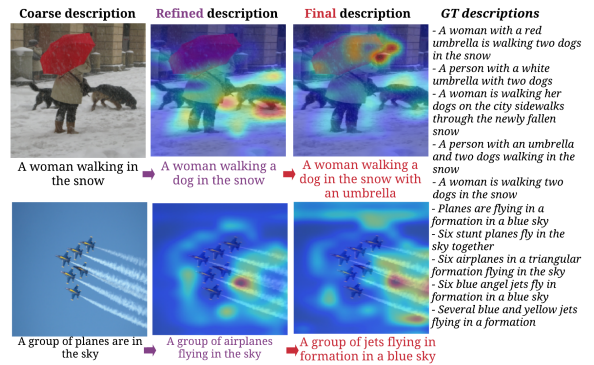

Figure 3: Visualizations of the generated captions and image attention maps on MSCOCO. Ground-Truth (**GT**) descriptions and the generated description of each stage are shown for each example. The columns from left to right correspond to the outputs of the three LSTM decoders from coarse to fine (coarse: *black*, refined: *purple*, final: *red*).

## Qualitative Analysis

To demonstrate that using the proposed coarse-to-fine approach can generate better image descriptions stage-by-stage that correlate well with the adaptively attended regions, we visualize the spatial attention weight for word in the generated captions. We upsample the attention weights by a factor of 16 and apply a Gaussian filter to make it the same size as the input image, and stack all the upsamped spatial attention maps into the original input image.

---
[2] https://competitions.codalab.org/competitions/3221

| Approach | BLEU-1 | BLEU-2 | BLEU-3 | BLEU-4 | METEOR | ROUGE-L | CIDEr | SPICE |
|---|---|---|---|---|---|---|---|---|
| Google NIC (Vinyals et al. 2015) | — | — | — | 27.7 | — | 23.7 | 85.5 | — |
| Hard-Attention (Xu et al. 2015) | 70.7 | 49.2 | 34.4 | 24.3 | 23.9 | — | — | — |
| Soft-Attention (Xu et al. 2015) | 71.8 | 50.4 | 35.7 | 25.0 | 23.0 | — | — | — |
| VAE (Pu et al. 2016) | 72.0 | 52.0 | 37.0 | 28.0 | 24.0 | — | 90.0 | — |
| Google NICv2 (Vinyals et al. 2016) | — | — | — | 32.1 | 25.7 | — | 99.8 | — |
| Attributes-CNN+RNN (Wu et al. 2016) | 74.0 | 56.0 | 42.0 | 31.0 | 26.0 | — | 94.0 | — |
| $CNN_\mathcal{L}$+RHN (Gu et al. 2017a) | 72.3 | 55.3 | 41.3 | 30.6 | 25.2 | — | 98.9 | 18.3 |
| PG-SPIDEr-TAG (Liu et al. 2017c) | 75.4 | 59.1 | 44.5 | 33.2 | 25.7 | 55.0 | 101.3 | — |
| Adaptive (Lu et al. 2017) | 74.2 | 58.0 | 43.9 | 33.2 | 26.6 | — | 108.5 | — |
| SCST:Att2in (Rennie et al. 2017) | — | — | — | 33.3 | 26.3 | 55.3 | 111.4 | — |
| SCST:Att2in (Ens. 4) (Rennie et al. 2017) | — | — | — | 34.8 | 26.9 | 56.3 | 115.2 | — |
| **Stack-Cap (C2F)** | **78.6** | **62.5** | **47.9** | **36.1** | **27.4** | **56.9** | **120.4** | **20.9** |

Table 3: Comparisons of the image captioning performance of the existing methods on MSCOCO Karpathy test split. Our Stack-Cap (C2F) model with the coarse-to-fine learning achieves significant gains across all metrics.

| Approach | BLEU-1 | | BLEU-2 | | BLEU-3 | | BLEU-4 | | METEOR | | ROUGE-L | | CIDEr | |
|---|---|---|---|---|---|---|---|---|---|---|---|---|---|---|
| | c5 | c40 | c5 | c40 | c5 | c40 | c5 | c40 | c5 | c40 | c5 | c40 | c5 | c40 |
| Google NIC | 71.3 | 89.5 | 54.2 | 80.2 | 40.7 | 69.4 | 30.9 | 58.7 | 25.4 | 34.6 | 53.0 | 68.2 | 94.3 | 94.6 |
| Hard-Attention | 70.5 | 88.1 | 52.8 | 77.9 | 38.3 | 65.8 | 27.7 | 53.7 | 24.1 | 32.2 | 51.6 | 65.4 | 86.5 | 89.3 |
| PG-SPIDEr-TAG | 75.1 | 91.6 | 59.1 | 84.2 | 44.5 | 73.8 | 33.1 | 62.4 | 25.5 | 33.9 | 55.1 | 69.4 | 104.2 | 107.1 |
| Adaptive | 74.8 | 92.0 | 58.4 | 84.5 | 44.4 | 74.4 | 33.6 | 63.7 | 26.4 | 35.9 | 55.0 | 70.5 | 104.2 | 105.9 |
| SCST:Att2in (Ens. 4) | 78.1 | 93.1 | 61.9 | 86.0 | 47.0 | 75.9 | 35.2 | 64.5 | 27.0 | 35.5 | 56.3 | 70.7 | 114.7 | 116.7 |
| Ours: Stack-Cap (C2F) | 77.8 | **93.2** | 61.6 | **86.1** | 46.8 | **76.0** | 34.9 | **64.6** | 27.0 | 35.6 | 56.2 | 70.6 | 114.8 | **118.3** |

Table 4: Leaderboard of the published image captioning models (as of 10/09/2017) on the online MSCOCO test server. Our single Stack-Cap model trained with the coarse-to-fine learning yields comparable performance with the state-of-the-art approaches on all reported metrics.

Figure 3 shows some generated captions. By reasoning via multiple attention layers progressively, the Stack-Cap model can gradually filter out noises and pinpoint the regions that are highly relevant to the current word prediction. We can find that our Stack-Cap model learns alignments that correspond strongly with human intuition. Taking the first image as an example, compared with the caption generated in the coarse stage, the first refined caption generated by the first fine decoder contains "*dog*," and the second fine decoder not only produces "*dog*," but also identifies "*umbrella*."

Besides, our approach can generate more descriptive sentences. For example, the attention visualizations of the *jets* image show that the Stack-Cap model can query the relationship of those "*jets*" as well as the *long trail of smoke* behind them, as there are strong attention weights that encompass this salient region. This, together with other examples, suggests that the stacked attention can more effectively explore the visual information for sequence prediction. In other words, our approach via the stacked attention can consider visual information in the image from coarse to fine, aligning well with the human visual system, where we usually use a coarse-to-fine procedure to understand pictures.

## Conclusion

In this paper, we have presented a coarse-to-fine image captioning model which utilizes a stacked visual attention model in conjunction with multiple LSTM networks to achieve better image descriptions. Unlike the conventional one-stage models, our approach allows generating captions from coarse to fine, which we found to be very beneficial for image captioning. Our model achieves comparable performance with the state-of-the-art approach using ensemble on the online MSCOCO test server. Future research directions include integrating extra attributes learning into image captioning, and incorporating beam search into the training procedure.

**Acknowledgements:** This research is supported by the National Research Foundation, Prime Ministers Office, Singapore, under its IDM Futures Funding Initiative, and NTU CoE Grant. This research was carried out at the ROSE Lab at the Nanyang Technological University, Singapore. The ROSE Lab is supported by the National Research Foundation, Prime Ministers Office, Singapore, under its IDM Futures Funding Initiative and administered by the Interactive and Digital Media Programme Office. We gratefully acknowledge the support of NVAITC (NVIDIA AI Tech Center) for our research at NTU ROSE Lab, Singapore.